\documentclass[10pt,twocolumn,letterpaper]{article}
\usepackage{iccv}              

\usepackage{bbding}
\usepackage{pifont}

\usepackage{epsfig}
\usepackage{amsmath}

\usepackage{amssymb}
\usepackage{bbm}
\usepackage{booktabs, multirow}

\usepackage{xcolor}

\graphicspath{{figures/}}

\usepackage{algorithm}
\usepackage{algorithmic}

\definecolor{iccvblue}{rgb}{0.21,0.49,0.74}
\usepackage[pagebackref,breaklinks,colorlinks,allcolors=iccvblue]{hyperref}

\title{WSCF-MVCC: Weakly-supervised Calibration-free Multi-view Crowd Counting}

\author{
Bin Li\thanks{Equal contribution.} \\
Shenzhen University\\
Shenzhen, China\\
{\tt\small libin2023@email.szu.edu.cn}
\and
Daijie Chen${}^*$\\
Shenzhen University\\
Shenzhen, China\\
{\tt\small chendaijie2022@email.szu.edu.cn}
\and
Qi Zhang\thanks{Corresponding author.}\\
Shenzhen University\\
Shenzhen, China\\
{\tt\small qi.zhang.opt@gmail.com}
}

\begin{document}

\maketitle

\begin{abstract}
Multi-view crowd counting can effectively mitigate occlusion issues that commonly arise in single-image crowd counting. Existing deep-learning multi-view crowd counting methods project different camera view images onto a common space to obtain ground-plane density maps, requiring abundant and costly crowd annotations and camera calibrations. Hence, calibration-free methods are proposed that do not require camera calibrations and scene-level crowd annotations. However, existing calibration-free methods still require expensive image-level crowd annotations for training the single-view counting module. Thus, in this paper, we propose a weakly-supervised calibration-free multi-view crowd counting method (WSCF-MVCC), directly using crowd count as supervision for the single-view counting module rather than density maps constructed from crowd annotations. Instead, a self-supervised ranking loss that leverages multi-scale priors is utilized to enhance the model's perceptual ability without additional annotation costs. What's more, the proposed model leverages semantic information to achieve a more accurate view matching and, consequently, a more precise scene-level crowd count estimation. The proposed method outperforms the state-of-the-art methods on three widely used multi-view counting datasets under weakly supervised settings, indicating that it is more suitable for practical deployment compared with calibrated methods. Code is released in \url{https://github.com/zqyq/Weakly-MVCC}.

\end{abstract}

\section{Introduction}
Single-image counting methods 
 \cite{li2018csrnet,jiang2020attention,Jiang2019Crowd}
have exhibited promising effectiveness in counting objects. However, when the scene becomes larger and the scene shape is irregular, the inherent limitations of existing single-image counting approaches hinder their practical application. For these wide-area scenes, recent researches use multiple camera views to estimate the crowd count of the whole scene, i.e., multi-view crowd counting (MVCC).

\begin{figure*}[t]
\small
\begin{center}
   \includegraphics[width=\linewidth]{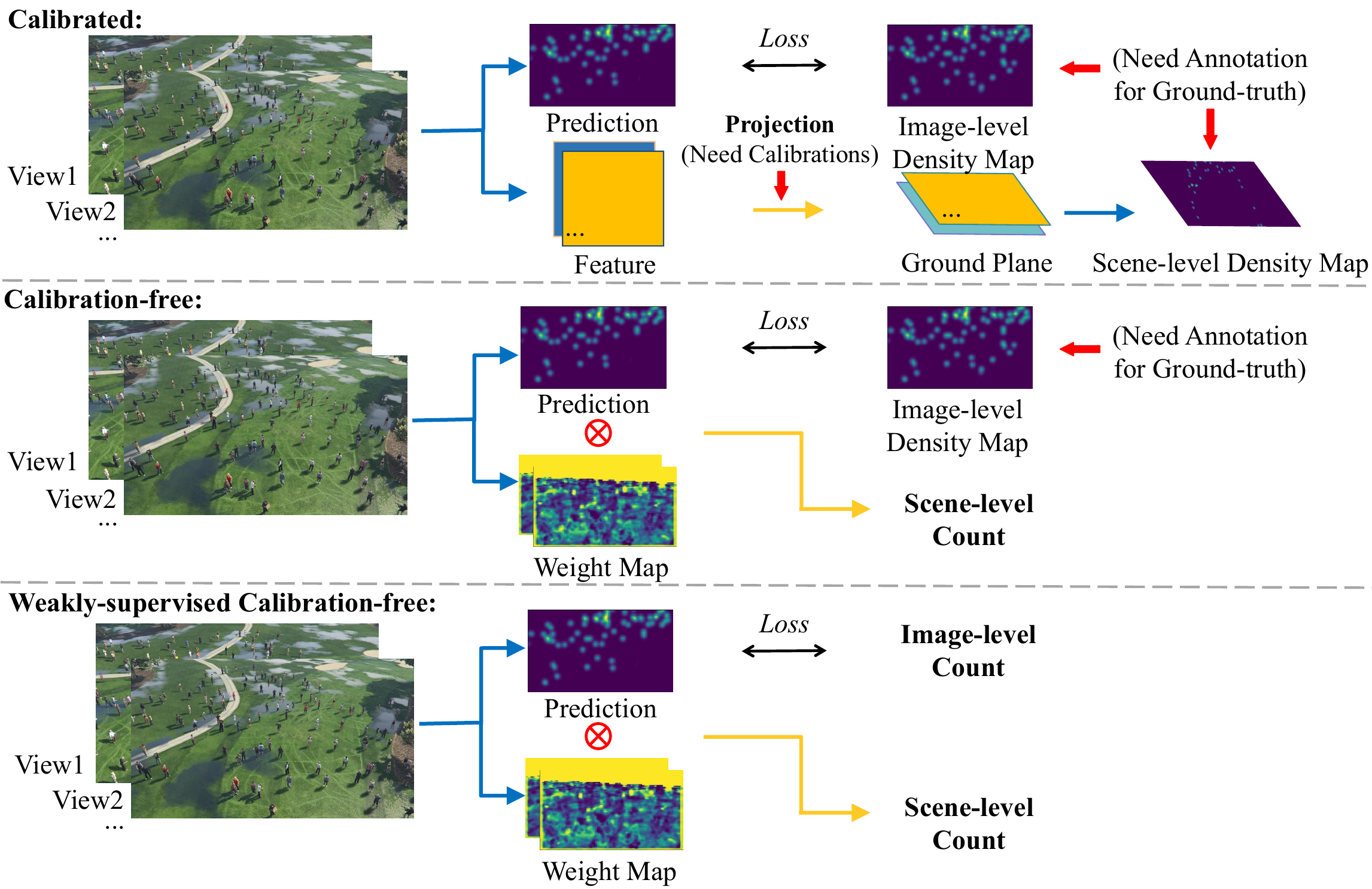}
\end{center}
\vspace{-0.7cm}
   \caption{
   Calibrated methods (top) require both camera calibration information to project image features and utilize annotations of image-level and scene-level people's locations.
   Calibration-free methods (middle) eliminate the need for camera calibration information and require the annotations of image-level people's locations. 
   Based on calibration-free methods, weakly-supervised calibration-free methods (bottom, \textbf{ours}) use annotations of image-level people counts rather than locations, significantly reducing annotation costs.
   }
\vspace{-0.4cm}
\label{fig:pipeline_different}
\end{figure*}

Generally, MVCC methods \cite{zhang2019wide,zhang2021CVCS,zhang2022wide,zhang2022single} make use of a common platform assumption to project image-level features onto the ground plane or a plane at a specific height. In the projection process, these methods rely on the camera calibration information (including intrinsic and extrinsic parameters). In addition, during the training phase, the model requires a large number of accurate human location annotations for supervision, as shown in Fig. \ref{fig:pipeline_different} top. All these factors limit the real application scenarios of MVCC. 

Thus, to extend and apply MVCC to a more practical scene, a calibration-free multi-view counting (CF-MVCC) method  \cite{zhang2022calibration} has been proposed, which aims to reduce the demand for camera calibration parameters and other annotated data, and achieves effective crowd count prediction results. 
It predicts a homography transformation matrix for aligning features from different camera views by using the same person's image coordinates in each view. To fuse image-level features from different perspectives, calibration-free MVCC methods directly predict the matching weight map for each camera view, rather than projecting features onto a common ground plane together. Finally, the scene-level crowd count can be calculated as a weighted sum over the predicted density maps, as shown in Fig. \ref{fig:pipeline_different} middle. However, the calibration-free MVCC method  \cite{zhang2022calibration} still relies on single-image crowd location annotations to train the single-view counting module, limiting its application scenarios. And it overly relies on the prediction accuracy of the homography matrix. When the homography matrix is inaccurately predicted, it leads to misalignment of features between the two camera views, making it difficult for the model to learn matching weight maps. Besides, the existing method ignores the critical semantic features between different camera pairs.

To address the issues mentioned above, we propose a weakly-supervised calibration-free multi-view crowd counting (WSCF-MVCC) method (see Fig. \ref{fig:pipeline_different} bottom). We directly use crowd count as a single-view module supervision rather than density maps constructed from crowd locations, reducing the annotation costs. Moreover, we utilize a self-supervised ranking loss to alleviate the influence of weakly supervised training. Furthermore, multi-view priors and semantic information are used to enhance the model's view-matching abilities for more accurate view matching.
Specifically, the proposed method estimates the total crowd count in the scene via 3 modules: 
1) The single-view crowd counting module (SVCC) consists of feature extraction and density map prediction submodules. Note that we also directly replace this module by using the existing single-image counting method as a fixed module, further reducing the model's demand for crowd annotation information. 
2) The matching weight estimation module (MWE) first estimates the homography between pairs of camera views. For each camera pair, the features from one camera view are then projected to the other view, concatenated, and used to estimate a matching probability map between the two camera views. Then, a weight map is calculated for each view using all the matching probability maps. Additionally, image content and distance information are used when calculating the weight maps to adjust for the confidence from each camera view. 
3) The multi-view counting estimation module (MVCE) calculates the total count as a weighted sum of the predicted single-view density maps using the estimated weight maps. In summary, the paper's contributions are as follows.
\begin{itemize}
\item We propose a weakly-supervised calibration-free multi-view crowd counting model, which significantly reduces annotation requirements compared with existing methods.
\item A self-supervised ranking loss leveraging multi-scale priors is proposed to enhance the model’s perceptual ability without additional annotation costs. Moreover, multi-view priors and semantic information are utilized to construct a view matching loss, achieving better view matching accuracy and model performance.
\item The proposed weakly-supervised method achieves better counting performance in three different datasets, indicating that the proposed method is more suitable for practical deployment.
\end{itemize}

\section{Related Work}

\textbf{Single-image crowd counting.}
Traditional single-image counting methods 
can be divided into three categories: detection-based 
 \cite{sabzmeydani2007detecting,wu2007detection}, 
regression-based 
 \cite{krizhevsky2012imagenet} 
or density map based methods  \cite{lempitsky2010learning}. 

Moreover, recent works  \cite{bai2020adaptive,liang2023crowdclip,peng2024single,guo2024regressor,lin2025pointtoregionlosssemisupervisedpointbased,yang2025tastemoretastebetter} based on deep neural networks (DNNs) have gradually become mainstream, focusing on density map estimation. 
 \cite{zhang2016single} introduced a multi-column CNN structure consisting of three columns of different receptive field sizes, which can enrich the features of people at different scales. 
 \cite{Kang2018Crowd} proposed to use the patch pyramid as input to extract multi-scale features. Besides, recent research has explored various forms of supervision, including regression methods and loss functions. 
Although these methods have significantly improved counting performance, they still rely on accurate annotations. However, this process is laborious and ambiguous due to heavy occlusion and scale variation. To extend the application scenarios of crowd counting, some researchers proposed the weakly supervised 
 \cite{liu2019exploiting,zhao2020active} 
or semi-supervised methods 
 \cite{sindagi2020learning}
. 
L2R \cite{liu2018leveraging} introduced a crop ranking loss by learning containment relationships to supervise unlabeled images.
Lin \textit{et al}  \cite{lin2022semi} proposed a new agency-guided semi-supervised counting approach, which associates the dense crowd areas of different images for obtaining reliable supervised signals.
However, a single camera view is insufficient to cover a large or extensive scene, and it is difficult to solve problems such as occlusion and scale transformation that exist in the image. Therefore, multi-view crowd counting methods have been proposed to 
solve these problems mentioned above.

\noindent
\textbf{Multi-view crowd counting.}
Previous \textbf{calibrated} multi-view crowd counting methods aim to predict the crowd count in a scene through images captured simultaneously by multiple different camera devices. 
Recently, \cite{zhang2019wide} proposed the first DNN-based multi-view counting method, which transforms each camera view's image-level features to the scene's ground plane and then fuses these features to estimate the scene-level density map. 
\cite{zhang2021CVCS} proposed a cross-view cross-scene multi-view counting model by camera selection and noise injection training. 
\cite{zhang2024multi} designed a view-contribution weight mechanism utilizing extra view-level supervision for better multi-view people detection under large scenes. 
MVOT \cite{zhang2024mahalanobis} proposed a point-supervised multi-view optimal transport loss for more accurate multi-view crowd localization.
However, previous multi-view crowd counting methods primarily project the camera views' features onto a common platform and require camera parameters for fusion and prediction. In this process, camera calibrations are needed in the testing stage, which limits their applicability in new scenarios where camera calibrations are unavailable. 
Hence, to reduce the model's demand for annotation data,  \cite{zhang2022calibration} proposes the first \textbf{calibration-free} counting methods. This method creates a matching weight map between each view by considering the number of cameras visible to a given pixel and the confidence level of each pixel. It calculates the total number of people in the scene with the matching weight maps and the predicted single-image density maps.
Although \cite{zhang2022calibration} effectively solves the dependence problem of multi-view crowd counting methods on camera calibration parameters, the fusion module’s blending weights are largely determined by the predicted homography matrix and need all the people's locations to generate the single-view density map, ignoring the critical semantic features and resulting in expensive annotation costs.

\section{Weakly-supervised Calibration-free MVCC}

The existing method  \cite{zhang2022calibration} effectively solves the dependence problem of multi-view crowd counting methods on camera calibration parameters. However, the fusion module’s blending weights are largely determined by the predicted homography matrix and require all people's locations to generate single-view density maps, thereby ignoring the critical semantic features and resulting in expensive annotation costs.

\begin{figure*}[t]
\begin{center}
   \includegraphics[width=\linewidth]{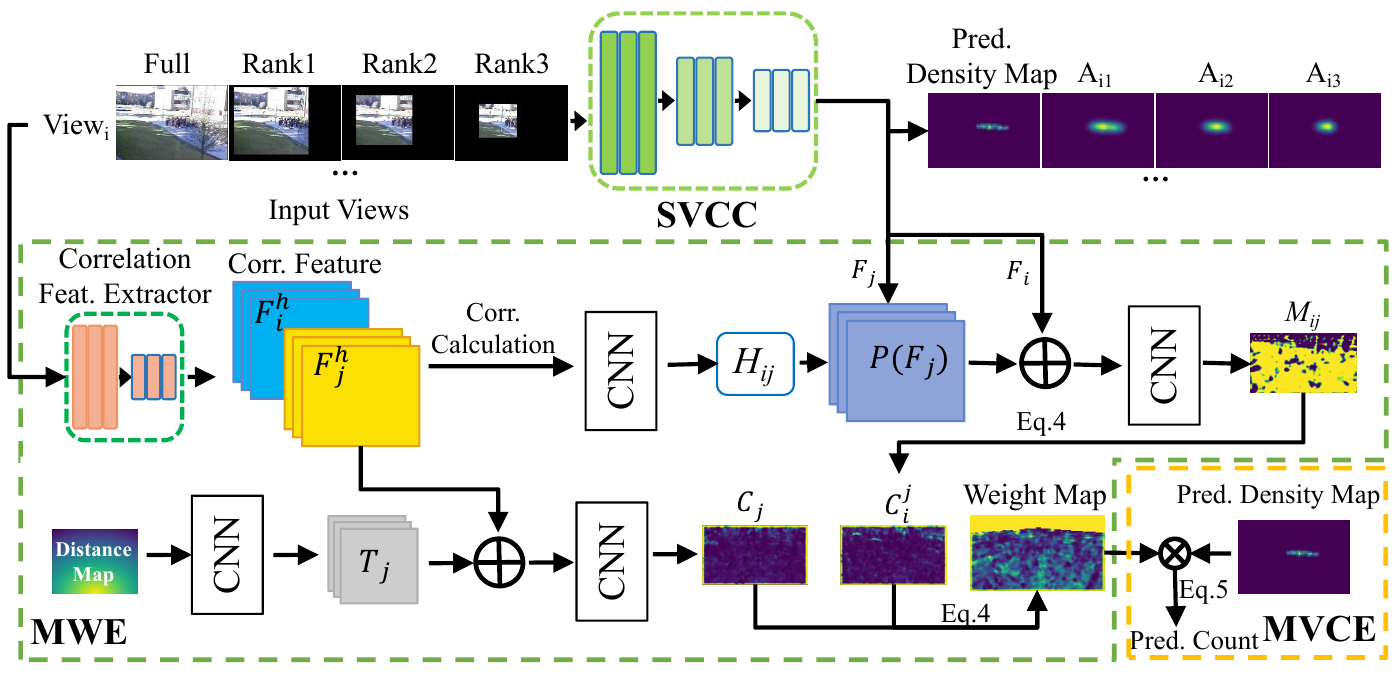}
\end{center}
\vspace{-0.5cm}
   \caption{The pipeline of the proposed WSCF-MVCC, including single-view crowd counting (SVCC), matching weight estimation (MWE), and multi-view crowd estimation (MVCE). Compared with CF-MVCC methods, we utilize the annotations of image-level people's counts to supervise the SVCC module rather than their locations.}
\vspace{-0.4cm}
\label{fig:pipeline}
\end{figure*}
To address the problems above, this section proposes a weakly-supervised calibration-free multi-view crowd counting method. We utilize the matching relationships between different views, require less crowd annotation information, and achieve good performance in multi-view crowd counting tasks. By exploring and effectively utilizing the matching relationship of viewpoints, the method can achieve counting performance comparable to those methods using more scene annotation information (including camera calibration information and more crowd head label information). Moreover, we utilize the number of all people as the ground truth (GT) of the single-view counting module to reduce annotation costs through weak supervision, and utilize multi-view prior and image semantic information to enhance the capabilities of view matching.

Specifically, our proposed model consists of the following three modules: single-view crowd counting, 
matching weight estimation, 
and multi-view counting estimation. 
The pipeline is illustrated in Fig. \ref{fig:pipeline}.


\subsection{Single-view Crowd Counting Module (SVCC)}
The SVCC module extracts the crowd feature from each camera view $i$ using a single-view feature extractor and predicts the crowd density map $D_i$ corresponding to that view through the feature decoder. 
To make a fair comparison with existing supervised crowd counting methods, we follow CVCS and use the first 7 layers of VGG-net as the feature extraction subnet, and the remaining layers of CSR-net as the decoder for predicting the density map $D_i$. 
The design of the loss function during the model training process is $l_{di}^{'} = \sum_{i=1}^{V}\left \| D_i - D_i^{gt} \right \|_{2}^{2} $, where $D_i$ and $D_i^{gt}$ are the predicted and ground-truth density maps respectively, and V is the number of camera views.
For weakly-supervised SVCC, since the function of the SVCC module is the same as that of the single-view crowd counting methods, we simultaneously explore the experimental results of using existing single-view crowd counting methods to directly replace the SVCC module. 
Specifically, we use CLTR  \cite{liang2022end} trained on the NWPU-Crowd dataset to replace the SVCC module for crowd prediction. 
Moreover, the predicted density map also focuses on the position information of the crowd in the scene, which can provide corresponding matching keypoints for the subsequent calculation of the perspective matching score. It is worth noting that we don't require additional scene crowd annotation data for retraining, which can further reduce the model's annotation demand. 
\begin{figure*}[t]
\begin{center}
   \includegraphics[width=\linewidth]{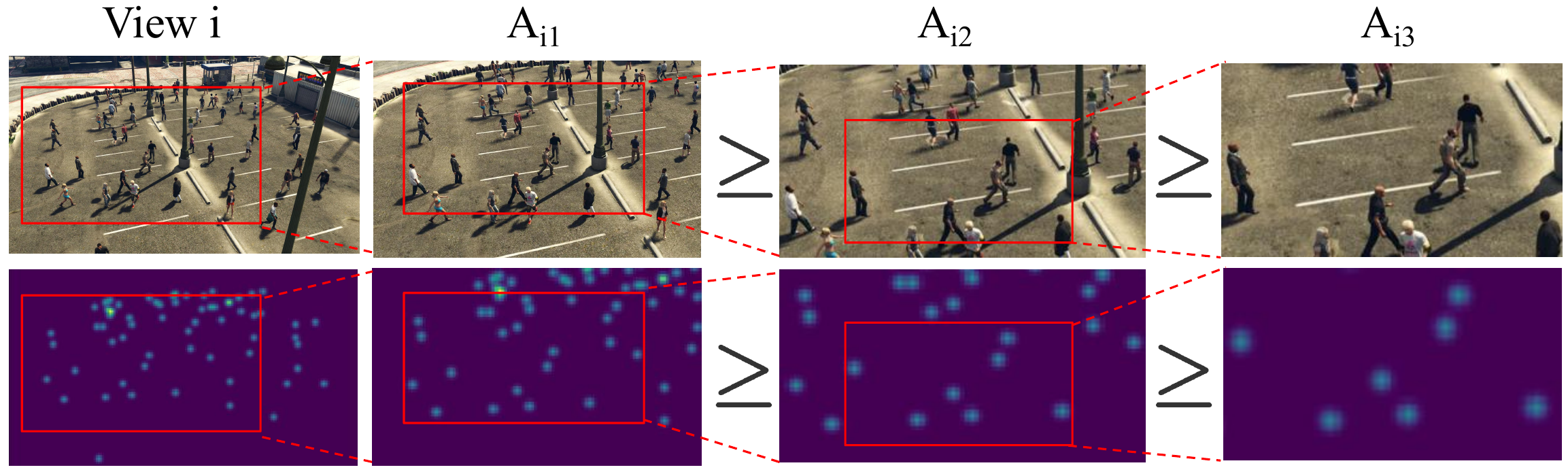}
\end{center}
\vspace{-0.6cm}
   \caption{Illustration of the ranking loss mechanism. The number of people contained in a certain area of the same image must be greater than or equal to the number of people contained in any sub-area within that area.}
\vspace{-0.4cm}
\label{fig:rank_loss}
\end{figure*}

Hence, in the SVCC module with weak supervision, we use CLTR and use the same feature extractor and transformer subnet as the feature extractor. Unlike CLTR, which directly predicts the number of crowd counting based on an extracted feature map, we use the layers of CSR-net as the decoder for predicting the crowd density map for each camera view. The sum of the predicted crowd density map can represent the number of people counted. Additionally, compared with  \cite{zhang2022calibration}, we only use the number of crowds as supervision rather than density maps constructed from crowd locations, which significantly reduces the annotation costs. Furthermore, 
through utilizing multi-scale priors, we adopt the common regional ranking loss to conduct weakly-supervised learning on the model during the fine-tuning process.
The principle of the loss function is shown in Fig. \ref{fig:rank_loss}. According to common sense, it is known that the number of people contained in a certain area of the same image must be greater than or equal to the number of people contained in any sub-area within that area. Therefore, there exists a constraint relationship that the prediction result on the sub-area must occur within the area. By adopting the regional loss function, the prediction results for local regions in different scenarios will be more reliable. In addition to supervising the prediction results of local areas, we also design a dedicated loss function to improve the performance of crowd counting. As in weakly-supervised crowd counting methods, we use the total count annotation information of the image to constrain the prediction result. In conclusion, the loss function can be written as:
\begin{scriptsize}
\begin{equation}
    l_{di} = \sum_{i=1}^{V}\left \| c_i - c_i^{gt} \right \|_{2}^{2} + 10*\sum_{i=1}^{V}\sum_{j=1}^{N}\sum_{k=j+1}^{N}max(C(A_{ik}) -C(A_{ij}), 0)
\end{equation}
\end{scriptsize}
where V represents the number of camera view, N indicates the number of sub-area on the same image, $c_i$ is the number of people prediction, $c_i^{gt}$ is the actual total number of people, $C(A_{ij})$ and $C(A_{ik})$ represent the number of people within subarea $A_{ij}$ and $A_{ik}$ in i-th view, and the visible area of $A_{ij}$ is more than $A_{ik}$, 
as shown in Fig. \ref{fig:pipeline}. 
In the experiment, we set $N=3$ and randomly select $A_{i1}$ from i-th view, then select $A_{i2}$ from $A_{i1}$, and finally select $A_{i3}$ from $A_{i2}$ for maintaining the effectiveness of the ranking loss (See Fig. \ref{fig:rank_loss}).

\subsection{Matching Weight Estimation Module (MWE)}
The core idea of the MWE module is to estimate the homography transformation matrix $H_{ij}$ between any 2 camera views $i$ and $j$. The corresponding relationship between camera pairs can be established using a homography matrix, and the matching score of different perspectives can be predicted based on the corresponding relationship, denoted as $M_{ij}$. 
After that, based on $M_{ij}$, we are to calculate the matching contribution weight of each pixel in a certain camera view toward the total number of people in the scene.
Specifically, we first use a CNN to estimate the homography transformation matrix from camera view $i$ to $j$. This CNN extracts the feature maps $F_i^h$ and $F_j^h$ from the two camera views. Secondly, the correlation map between $F_i^h$ and $F_j^h$ is computed, and a decoder is utilized to predict $H_{ij}$. For supervision, the ground-truth homography matrix $H_{ij}^{gt}$ is calculated based on the corresponding people's head locations in the two camera views. In particular, when the camera view pair has no overlapping field of view, a dummy homography matrix is used as the GT to indicate that the two camera views are non-overlapping. The loss function for training the homography estimation CNN is defined as:
\begin{small}
\begin{equation}
    l_h = \sum_{i=1}^V\sum_{j=1,j\ne i}^V\left \| H_{ij} - H_{ij}^{gt} \right \|_{2}^{2}
\end{equation}
\end{small}

After that, we can align the features of any two perspectives through the homography transformation matrix and learn the correlations between the features. We use a sub-network to predict the matching score map $M_{ij}$, whose elements indicate the probability of whether the given pixel in view $i$ has a match anywhere in view $j$. The input into the subnet is the concatenation of features $F_i^c$ extracted from view $i$, and the aligned features from view $j$, $P(F_j^c, H_{ij})$, where $P$ is the projection layer adopted from STN  \cite{Jaderberg2015Spatial}.
Note that, in a multiple camera-view scenario, the corresponding people's head locations contain extremely rich and crucial semantic information. In the set of corresponding head locations, any pair of position points can indicate that the corresponding positions in different views have a direct mapping relationship in space. Moreover, when counting the number of people in the scene across different views, the matching weight of the corresponding positions in different views should sum up to one. By leveraging the above-mentioned prior information, the model can be effectively assisted in better allocating the weight matching score between camera views. Therefore, we use the corresponding people's head labels to generate the ground truth $M_{ij}^{gt}$ for supervising the prediction of the view matching weight map. The loss used to train the matching score map estimation CNN is 
\begin{equation}
    l_d = \sum_{i=1}^V \sum_{j=1,j\ne i}^V \left\| M_{ij}*M_{ij}^{gt} - M_{ij}^{gt}\right\|
\end{equation}
where $M_{ij}^{gt}$ is a binarized graph.  The purpose of binarization is to enable the model to allocate weights reasonably only for the match of people's semantic information in the scene, and the weight values of the remaining areas will be learned by the network.

Finally, the confidence score map $C_i$ for camera view $i$ is generated from the image features and pixel-wise distance information. The camera view weight maps are computed as:
    \begin{equation}
        W_i=C_i/(C_i+\sum_{j\ne i}^V{C_j^i \odot M_{ij})}
    \end{equation}
where $\odot$ is element-wise multiplication, and $C_j^i = P(C_j, H_{ij})$ is the projection of confidence map $C_j$ to camera view $i$. Note that the views with higher confidence will have a higher contribution to the count of the corresponding pixel.

The confidence map $C_i$ is predicted by a CNN whose inputs are the image feature map $F_i^h$ and distance feature map $T_i$. Generally, $T_i$ is computed by feeding a distance map \cite{zhang2022calibration}, where each pixel represents the distance to the camera in the 3D scene, into a small CNN.

\begin{table*}[t]
\centering
\caption{Scene-level counting performance on the synthetic multi-scene dataset CVCS.}
\begin{tabular}
{l|l|l|cc}
\hline
      & Supervision & Method & MAE $\downarrow$ & NAE $\downarrow$ \\
\hline
    \multirow{4}{*}{Calibrated} 
    & \multirow{4}{*}{Fully-supervised}
    & CVCS\_backbone \cite{zhang2021CVCS}& 14.13 & 0.115  \\
    & & CVCS(MVMS) \cite{zhang2021CVCS,zhang2019wide} & 9.30 & 0.080  \\
    & & CVCS \cite{zhang2021CVCS} & 7.22 & 0.062 \\
    & & CountFormer \cite{mo2024countformer} & \textbf{4.79} & \textbf{0.039} \\
    
\hline
    \multirow{5}{*}{Calibration-free} 
    & \multirow{5}{*}{Fully-supervised}
    & Dmap\_weightedH \cite{zhang2022calibration} & 28.28 & 0.239  \\
    & & Dmap\_weightedA \cite{zhang2022calibration} & 19.85 & 0.165  \\
    & & Total\_count \cite{zhang2022calibration} & 18.89 & 0.157 \\
    & & CF-MVCC \cite{zhang2022calibration}& 13.90 & 0.118 \\
    & & WSCF-MVCC (Ours) & \textbf{12.99} & \textbf{0.111} \\
\hline
    \multirow{2}{*}{Calibration-free} 
    & \multirow{2}{*}{Weakly-supervised}
    & CF-MVCC-S \cite{zhang2022calibration}& 15.29 & 0.127  \\
    & & WSCF-MVCC-S (Ours) & \textbf{13.66} & \textbf{0.114}  \\
\hline
\end{tabular}
\vspace{-0.3cm}
\label{table:CVCS_counting_results}
\end{table*}

\subsection{Multi-view Counting Estimation Module (MVCE)}
Through the computed weight map $W_i$ for each view $i$, the total crowd count $S$ is the weighted sum of the density map predictions $D_i$. The calculation formula of $S$ is as follows:
    \begin{equation}
        S=\sum_{i=1}^V sum(W_i\odot D_i)
    \end{equation}
where 
sum is the summation over the map. During training, the total count loss is the MSE of the count prediction: $l_s=\left\| S-S^{gt} \right\|,$ where $S^{gt}$ is the GT. Finally, the loss for training the whole model can be written as:
    \begin{equation}
        Loss = l_s+\lambda l_{di}+\beta l_{d}+\gamma l_h
    \end{equation}
where $\lambda$, $\beta$, and $\gamma$ are hyperparameters to control the weight of these loss items.

\begin{table*}[t]
\small
\centering
\caption{Scene-level counting performance on real single-scene datasets.}
\begin{tabular}
{l|l|l|cc|cc}
\hline
      & \multirow{2}{*}{Supervision}& \multirow{2}{*}{Method} & \multicolumn{2}{c|}{CityStreet} & \multicolumn{2}{c}{PETS2009} \\
      & & & MAE $\downarrow$ & NAE $\downarrow$ & MAE $\downarrow$ & NAE $\downarrow$ \\
\hline
    \multirow{4}{*}{Calibrated} & \multirow{4}{*}{Fully-supervised} & MVMS \cite{zhang2019wide} & 8.01 & 0.096 & 3.49  & 0.124 \\
    & & 3D\_counting \cite{zhang20203d} & 7.54 & 0.091 & 3.15 & 0.113 \\
    & & CVF \cite{Zheng2021Learning} & 7.08 & - & 3.08 & - \\
    & & CountFormer \cite{mo2024countformer} & \textbf{4.72} & \textbf{0.058} & \textbf{0.74} & \textbf{0.030} \\
\hline
    \multirow{5}{*}{Calibration-free} & \multirow{5}{*}{Fully-supervised}
    & Dmap\_weightedH \cite{zhang2022calibration} & 9.84 & 0.107 & 4.23 & 0.136 \\
    & & Dmap\_weightedA \cite{zhang2022calibration}& 9.40 & 0.123 & 6.25 & 0.252 \\
    & & Total\_count \cite{zhang2022calibration}& 11.28 & 0.152 & 6.95 & 0.265 \\
    & & CF-MVCC \cite{zhang2022calibration} & 8.06 & 0.102 & 3.46 & 0.116 \\
    & & WSCF-MVCC (Ours)& \textbf{7.28} & \textbf{0.093} & \textbf{2.92} & \textbf{0.103} \\
\hline
    \multirow{2}{*}{Calibration-free} & \multirow{2}{*}{Weakly-supervised}
    & CF-MVCC-S \cite{zhang2022calibration}& 7.74 & 0.110 & 3.49 & 0.116 \\
    & & WSCF-MVCC-S (Ours)& \textbf{7.40} & \textbf{0.102} & \textbf{3.26} & \textbf{0.110} \\
\hline
\end{tabular}
\label{table:real_datasets_counting_results}
\end{table*}

\begin{table}[t]
\centering
\caption{Ablation study on whether using distance and semantic information in the matching weight map estimation.}
\vspace{-0.2cm}
\begin{tabular}
{l|cc}
\hline
    Method & MAE $\downarrow$ & NAE $\downarrow$ \\
\hline
    Base & 16.13 & 0.139 \\
    Base+Dist. & 13.90 & 0.118 \\
    Base+Dist.+Seg. & 12.99 & 0.111 \\
\hline
\end{tabular}
\vspace{-0.3cm}
\label{table:ablation_annotation_results}
\end{table}

\section{Experiments}
\subsection{Experiment Settings}

\textbf{Ground-truth.} 
We use the single-view density map, the homography transformation matrix, and the scene crowd count as ground-truth for training. In the SVCC module, we also explore using a single-view crowd count to replace the single-view density map for training the model. The GT for the single-view density maps is constructed in a manner similar to typical single-image counting methods. The ground-truth homography transformation matrix of a camera-view pair is calculated with the corresponding people's head coordinates (normalized to [-1, 1]). If there are no common people in the two camera views, we define $H=[0,0,-10;0,0,-10;0,0,1]$ as a `dummy' homography matrix that represents the situation of no overlapped region. As for the ground-truth single-view crowd count or the ground-truth people count, we only require the number of the single-view level or the total scene-level, respectively. Compared with the previous multi-view crowd counting methods  \cite{zhang2022calibration}, our approach reduces the reliance on camera calibration information, image-level crowd annotations, and scene-level crowd annotations. 

\textbf{Training and evaluation.} 
The training process of this method is carried out in stages. First, we train the SVCC and the homography estimation module respectively. Then fix both of them and train the remaining MWE and MVCE modules. On the large synthetic dataset, we use learning rates of $10^{-3}$. On the real scene datasets, the learning rate is $10^{-4}$. Mean absolute error (MAE) and mean normalized absolute error (NAE) of the predicted counts are used as the evaluation metrics.

\textbf{Datasets.} 
We validate the proposed calibration-free multi-view counting on both a synthetic dataset, CVCS \cite{zhang2021CVCS}, and real datasets, CityStreet \cite{zhang2019wide} and PETS2009 \cite{ferryman2009pets2009}. CVCS is a cross-scene synthetic dataset for the multi-view counting task. Each scene contains 100 frames (with an input image resolution of 640×360) and about 100 camera views (280k total images). CityStreet, PETS2009 are two real single-scene datasets for multi-view counting. CityStreet contains 3 camera views and 300 multi-view frames (676×380) for training and 200 for testing. PETS2009 contains 3 camera views and 1105 multi-view frames (384×288) for training and 794 for testing.

\subsection{Experiment Results}
\subsubsection{Weakly-supervised counting performance.} 
We compare the scene-level counting performance of the existing methods on CVCS \cite{zhang2021CVCS}, CityStreet \cite{zhang2019wide}, and PETS2009 \cite{ferryman2009pets2009}. In Table \ref{table:CVCS_counting_results}, we show the experiment results on the CVCS dataset, and the proposed WSCF-MVCC method achieves the best performance among the calibration-free methods. 
Compared with comparisons Dmap\_weightedH and Dmap\_weighteddA, which only consider the geometric constraints of the camera or the image content, the performance of these methods is worse than that of CF-MVCC \cite{zhang2022calibration}. 

Total\_count replaces the supervision of the crowd density map in CVCS with total count loss, so the prediction results will be disturbed by the changes in the scene. 

`-S' represents using the annotations of the crowd number for training the SVCC rather than the density map constructed from crowd locations.
Compared with CF-MVCC(-S), WSCF-MVCC(-S) can further enhance the model's performance after supervised learning of the view matching score map, which shows the same advantages in Table \ref{table:real_datasets_counting_results}. Finally, the performance of WSCF-MVCC is inferior to that of the calibration methods; however, it still narrows the gap with the calibrated methods and is superior to the CVCS backbone method. It can be demonstrated that the method proposed in this study can be used for scene-level crowd count prediction without using any calibration parameters.

In Table \ref{table:real_datasets_counting_results}, we evaluate all the calibration-free methods on two real single-scene datasets.
According to the experiment results on real datasets, our proposed method achieves better performance than the other calibration-free methods.

Additionally, the performance of WSCF-MVCC is comparable to the calibrated multi-view crowd counting methods MVMS \cite{zhang2019wide} and 3DCount \cite{zhang20203d}, and slightly worse than CVF \cite{Zheng2021Learning}. The reason for this result might lie in the fact that calibration-free methods can implicitly learn specific camera geometric parameters during the fusion step. 

Compared with the experimental results on the CVCS dataset, the experimental results on the two real-world datasets are closer to those of calibrated multi-view crowd counting methods. 

As a cross-scene dataset, CVCS has a test set covering diverse scene types and complex camera distributions, which significantly increases the difficulty of predicting transformation relationships between different camera perspectives in new scenes. However, the experimental results in Table \ref{table:CVCS_counting_results} demonstrate that WSCF-MVCC can effectively predict transformation relationships in different scenes and estimate the number of people in the scene based on these relationships.

\subsection{Visualization Results}
\begin{figure*}[t]
\small
\begin{center}
   \includegraphics[width=1\linewidth]{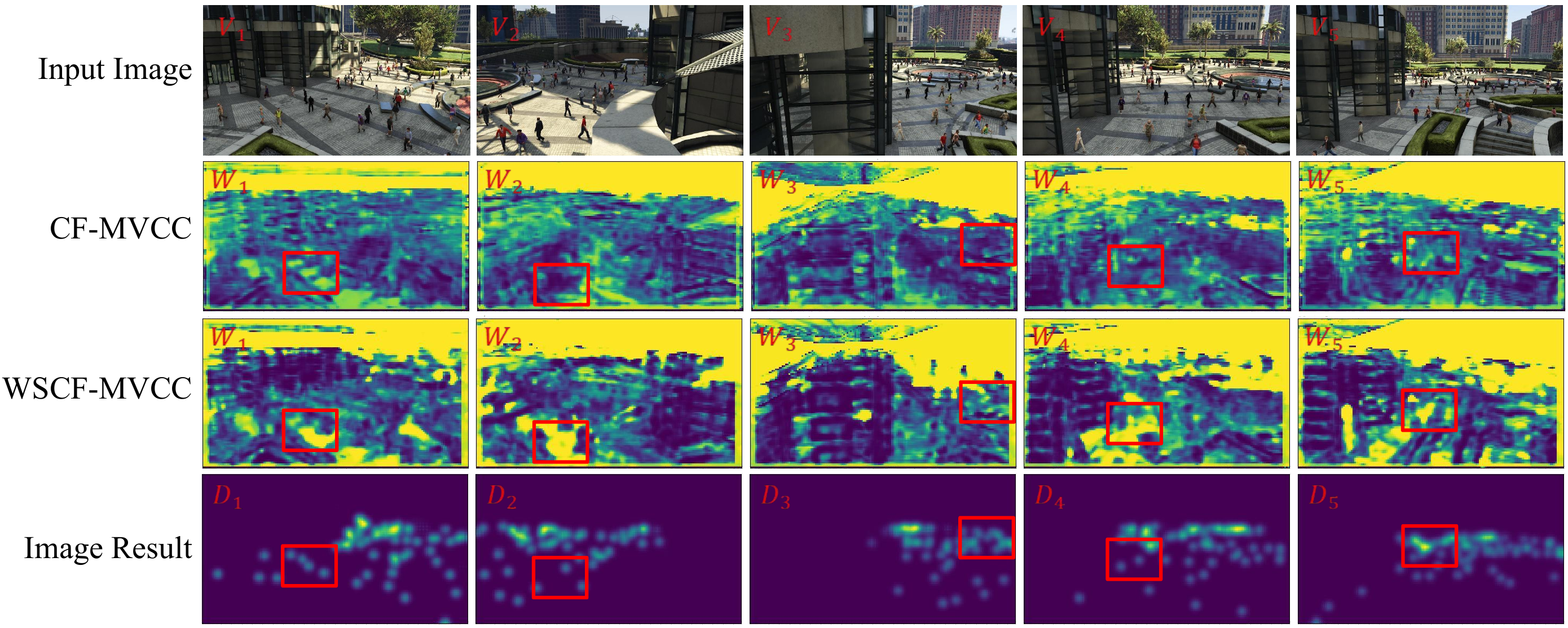}
\end{center}
\vspace{-0.6cm}
   \caption{Visualization results of weight maps $W$ and density maps $D$.
   The red boxes indicate that our MSCF-MVCC method can obtain a more reliable match weight.
   }
\label{fig:result1}
\end{figure*}

\begin{figure*}[t]
\small
\begin{center}
   \includegraphics[width=\linewidth]{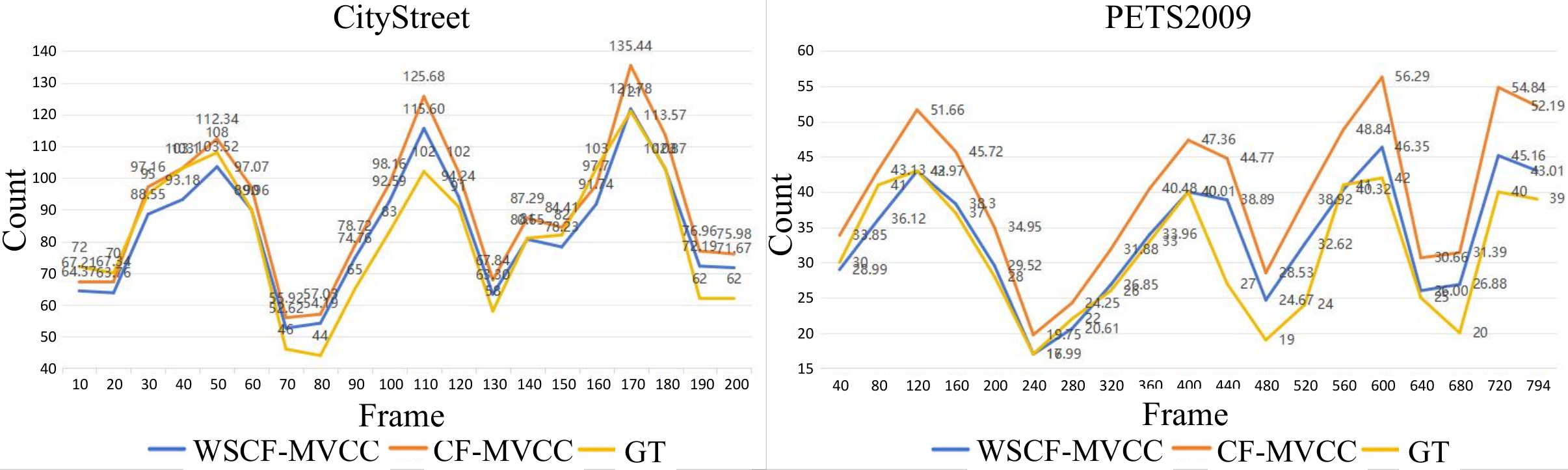}
\end{center}
 \vspace{-0.8cm}
   \caption{The counting results on CityStreet and PETS2009. It shows that the results of the proposed WSCF-MVCC are closer to the GT.}
\label{fig:result3}
\end{figure*}

We show the visualization results of weight maps $W$ and density maps $D$ in Fig. \ref{fig:result1}. 
The red boxes indicate regions with a more reliable match weight, showing that the proposed WSCF-MVCC possesses enhanced view-matching capabilities.
Additionally, we present the experiment results on CityStreet and PETS2009 datasets. In Fig. \ref{fig:result3}, the results of WSCF-MVCC are closer to the ground-truth.

\subsection{Ablation Studies} 
In this section, we conduct various ablation studies on the CVCS dataset.

\begin{table}[t]
\small
\centering
\caption{Ablation study on single-view counting networks for SVCC module.}
\vspace{-0.2cm}
\begin{tabular}
{cl|cc}
\hline
     & SVCC Method & MAE $\downarrow$ & NAE $\downarrow$ \\
\hline
    \multirow{3}{*}{Weakly-supervised} 
    & CSRNet & 15.29 & 0.128  \\
    & DM-Count & 16.49 & 0.139   \\
    & CLTR & 13.66 & 0.114  \\
    
\hline
\end{tabular}
\vspace{-0.3cm}
\label{table:ablation_svcc_results}
\end{table} 

\noindent
\textbf{Ablation study on matching weight map.} 
We conduct an ablation study to show which information can help the model to predict a more reliable matching weight map. First, we conduct a basic method that utilizes only image features to estimate a matching map without any other additional information, i.e., `Base'. Next, we add distance information, denoted as `Base+Dist.'. Finally, we add both distance and semantic information as `Base+Dist.+Seg.', i.e., WSCF-MVCC. The results are shown in Table \ref{table:ablation_annotation_results}. Compared with not using additional information, using distance information can improve prediction performance. Furthermore, using both distance and semantic information can further enhance model performance. This demonstrates that with effective information supervision, matching weight maps can effectively predict the weight relationships between different perspectives.

\begin{table}[t]
\centering
\caption{Ablation study on the variable testing camera number (3, 5, 7, 9) of the proposed method on the CVCS dataset, which is trained on 5 camera views.}
\begin{tabular}
{c@{\hspace{0.08cm}}c@{\hspace{0.08cm}}|c@{\hspace{0.08cm}}c@{\hspace{0.08cm}}|c@{\hspace{0.08cm}}c@{\hspace{0.08cm}}}
\hline
        & \multirow{2}{*}{Number} & \multicolumn{2}{c|}{CF-MVCC} & \multicolumn{2}{c}{WSCF-MVCC} \\
        & & MAE $\downarrow$ & NAE $\downarrow$ & MAE $\downarrow$ & NAE $\downarrow$ \\
\hline
    & 3 & 11.01 & 0.107 & 14.43 & 0.137   \\
    & 5 & 13.90 & 0.118 & 12.99 & 0.111  \\
    & 7 & 18.45 & 0.147 & 15.77  & 0.128 \\
    & 9 & 22.23 & 0.174 & 21.89 & 0.173  \\ 
\hline
\end{tabular}
\label{table:ablation_number_results}
\end{table}

\noindent
\textbf{Ablation study on single-view counting network.} 
We explore using a weakly supervised approach to train the existing single-view counting network as the SVCC module, thereby reducing the model's demand for single-view annotation data. In Table \ref{table:ablation_svcc_results}, we consider 3 commonly used single-view counting modules  \cite{li2018csrnet,wang2020distribution,liang2022end} as the SVCC module and evaluate it on the CVCS dataset. The results demonstrate that CLTR  \cite{liang2022end} achieves better performance than using other single-view counting networks in the SVCC module. Thus, we use it in all experiments.

\noindent
\textbf{Ablation study on the view number.} 
The third ablation study is conducted on the variable camera views during the testing stage. To show the influence of the number of camera views, we test WSCF-MVCC with different numbers of input cameras, where the model is trained on the CVCS dataset using 5 camera views as input. The results are presented in Table \ref{table:ablation_number_results}. WSCF-MVCC achieves a more robust performance than the calibration-free method CF-MVCC. When the number of test camera views equals or exceeds the number of training views, as in the CF-MVCC method, the results decrease as the number of cameras increases. The reason is that the error in weight map prediction might increase when the number of camera views changes, but this increase can be constrained by the supervision of useful semantic information. In addition, when the number of test cameras is fewer than the training views, the performance also degrades. The possible reason is that there is less effective matching information provided between perspectives.

\section{Conclusion}
In this paper, we focus on the weakly-supervised calibration-free multi-view crowd counting task (WSCF-MVCC). 
Existing calibration-free methods overlook critical semantic features and require expensive annotation costs to construct supervision for training the single-view module.
Thus, we directly use crowd count as supervision for the single-view module, significantly reducing annotation costs. With multi-scale priors, we adopt a weakly-supervised ranking loss to enhance the model's perception ability without extra annotation costs, mitigating the limitations of weak supervision. Moreover, we design a view matching loss that uses multi-view priors and semantic information, enabling the model to achieve better view matching ability and performance. Extensive experiments demonstrate that the proposed WSCF-MVCC is more suitable for practical deployment compared with calibrated MVCC methods and CF-MVCC methods. 
Additionally, the proposed losses possess good generalization abilities and can be applied to existing calibration-free single-view or multi-view methods 
to enhance the model's performance.

\section*{Acknowledgements}
This work was partially supported by ‌the National Natural Science Foundation of China (NSFC) 62202312, DEGP Innovation Team (2022KCXTD025), Scientific Foundation for Youth Scholars of Shenzhen University, Shenzhen University Teaching Reform Key Program (JG2024018), and Scientific Development Funds from Shenzhen University.

\bibliographystyle{ieeenat_fullname}
\bibliography{egbib}

\end{document}